\documentclass[runningheads]{llncs}
\usepackage[T1]{fontenc}
\usepackage{paralist}
\usepackage{hyperref}
\usepackage{algorithm}
\usepackage{algorithm}
\usepackage{balance}
\usepackage{amsmath}
\usepackage{amssymb}
\usepackage{multirow}
\usepackage{mathtools}
\usepackage{algpseudocode}
\usepackage{textcomp}
\usepackage{gensymb}
\usepackage{algorithm}
\usepackage{graphicx}
\usepackage{subcaption}
\usepackage{mwe}
\usepackage[normalem]{ulem}
\usepackage{color}
\usepackage{soul}
\usepackage{ragged2e}

\usepackage{lipsum}

\usepackage{forloop}
\newcounter{ct}

\definecolor{darkblue}{RGB}{0,0,120}
\definecolor{urlblue}{RGB}{32,43,255}
\definecolor{persoRed}{RGB}{175,35,24}
\usepackage{paralist}

\begin{document}
%
% \title{Interpretable Counterfactual Explanations for Time Series Data Guided by Shapelets}
\title{Motif-guided Time Series Counterfactual Explanations}

\titlerunning{Motif-guided Counterfactual Explanations of Time Series Data}
\author{Peiyu Li\inst{1} \and
Souka\"ina Filali Boubrahimi\inst{1} \and
Shah Muhammad Hamdi\inst{2}}
\authorrunning{P. Li, S. Boubrahimi et al.}
\institute{Utah State University,
 Logan, UT 84322, USA \and
 New Mexico State University,
 Las Cruces, NM 88003, USA
}
\maketitle              % typeset the header of the contribution
\begin{abstract}
% The abstract should briefly summarize the contents of the paper in 150--250 words.
With the rising need of interpretable machine learning methods, there is a necessity for a rise in human effort to provide diverse explanations of the influencing factors of the model decisions. To improve the trust and transparency of AI-based systems, the EXplainable Artificial Intelligence (XAI) field has emerged. The XAI paradigm is bifurcated into two main categories: feature attribution and counterfactual explanation methods. While feature attribution methods are based on explaining the reason behind a model decision, counterfactual explanation methods discover the smallest input changes that will result in a different decision. In this paper, we aim at building trust and transparency in time series models by using motifs to generate counterfactual explanations. We propose Motif-Guided Counterfactual Explanation (MG-CF), a novel model that generates intuitive post-hoc counterfactual explanations that make full use of important motifs to provide interpretive information in decision-making processes. To the best of our knowledge, this is the first effort that leverages motifs to guide the counterfactual explanation generation. We validated our model using five real-world time-series datasets from the UCR repository. Our experimental results show the superiority of MG-CF in balancing all the desirable counterfactual explanations properties in comparison with other competing state-of-the-art baselines.

\end{abstract}
\section{Introduction} 
Recently, time-series data mining has played an important role in various real-life applications, such as healthcare \cite{lin2005approximations}, astronomy \cite{boubrahimi2018spatiotemporal}, and 
aerospace \cite{lin2004visually}. The availability of big 
data provides researchers the unprecedented opportunity to deploy high accurate models in real-life applications. The success of deep learning time series models has been validated by their high accuracy. However, the black-box nature makes the decision-making process of high accuracy models less explained \cite{adadi2018peeking}. This represents a big barrier for decision makers in trusting the system. For example, in life-changing decisions such as critical disease management, it is important to know the reasons behind the critical decision of choosing a treatment line over another one \cite{kundu2021ai}. An inappropriate treatment decision can cost human lives in addition to a substantial monetary loss \cite{xu2019deep}. Therefore, it is vital to understand the opaque models' decision process by either prioritizing interpretability during the model development phase, or developing post-hoc explanation solutions. 

\par To build trust between humans and decision-making systems, EXplainable Artificial Intelligence (XAI) methods are increasingly accepted as effective tools to explain machine learning models' decision-making processes. XAI methods aim at increasing the interpretability of the models whilst maintaining the high-performance levels. There are two XAI dominant paradigms: intrinsic interpretable and post-hoc explanations for opaque models \cite{zhou2021s}. Examples of intrinsic interpretable models include linear regression, decision trees, rule sets, etc. On the other hand, examples of post-hoc explanation methods include LIME \cite{ribeiro2016should}, native guide, and SHAP \cite{slack2019can} \cite{verma2020counterfactual}. In this paper, we focus only on post-hoc XAI methods. A lot of efforts have been made to provide post-hoc XAI for image and vector-represented data while significantly less attention has been paid to time series data \cite{nguyen2020model}. The complex and time-evolving nature of time series makes the explanation models one of the most challenging tasks \cite{delaney2021instance}.

On the light of desirable counterfactual method properties, we propose a Motif-Guided Counterfactual Explanation (MG-CF), a novel model that generates interpretable, intuitive post-hoc counterfactual explanations of time series classification predictions that balance between validity, proximity, interpretability, contiguity and efficiency. The contributions are the following:

\begin{compactenum}
    \item We incorporate motif mining algorithms for guiding the counterfactual search. In particular, we make full use of the interpretability power of motifs to provide interpretable information in the decision-making process. Our method does not require the use of class activation maps to search for the counterfactual explanation, which makes it model-agnostic. 
    \item We conduct experiments on six real-life datasets from various domains (image, spectro, ECG, and motion) and show the superiority of our methods over state-of-the-art models. 
\end{compactenum}

To the best of our knowledge, this is the first effort to leverage the a-priori mined motifs to produce counterfactual explanations for time series classification. The rest of this paper is organized as follows: in section 2, background and related works are described. Section 3 introduces the preliminary concepts. Section 4 describes our proposed method in detail. We present the experimental results and evaluations in comparison to other baselines in section 5. Finally, we conclude our work in section 6.

\section{Background and Related work}
In the post-hoc interpretability paradigm, one of the most prominent approaches is to identify important features given a prediction through local approximation, such as LIME \cite{ribeiro2016should}, LORE \cite{guidotti2018local}, TS-MULE \cite{schlegel2021ts}, and SHAP \cite{lundberg2017unified}.
In particular, LIME is a feature-based approach that operates by fitting an interpretable linear model to the classifier’s predictions of random data samples, weighted based on their distance to the test sample. LORE is an extension work based on LIME, which is a local black box model-agnostic explanation approach based on logic rules. TS-MULE is also an extension to LIME with novel time series segmentation techniques and replacement methods to enforce a better non-informed values exchange. SHAP is a unified framework that operates by calculating feature importance values using model parameters. In addition, visualizing the decision of a model is also a common technique for explaining model predictions \cite{mothilal2020explaining}. In the computer vision domain, visualization techniques have been widely applied to different applications successfully, such as highlighting the most important parts of images to class activation maps (CAM) in convolutional neural networks \cite{mahendran2015understanding}. Schlegel et al. \cite{schlegel2019towards} tested the informativeness and robustness of different feature-importance techniques in time series classification. LIME was found to produce poor results because of  the large dimensionality by converting time to features; in contrast, saliency-based approaches and SHAP were found to be more robust across different architectures

An alternative method to feature-based methods has been proposed by Wachter et al. \cite{wachter2017counterfactual}, which aims at minimizing a loss function and using adaptive Nelder-Mead optimization to encourage the counterfactual to change the decision outcome and keep the minimum Manhattan distance from the original input instance. Similarly, another method that is used to deal with the plausibility and feasibility issues of the generated counterfactual explanation has been proposed, which is called GeCo. The model achieves the desirable counterfactual properties by introducing a new plausibility-feasibility language (PLAF) \cite{schleich2021geco}. Both GeCo and wCF focus on structured tabular datasets. Since these two methods explore a complete search space, they are not adequate to be used in a high dimensional feature space such as in the case of time-series data.

Recently, an instance-based counterfactual explanation for time series classification has been proposed \cite{delaney2021instance}. The instance-based counterfactual explanation uses the explanation weight vector (from the Class Activation Mapping) and the in-sample counterfactual (NUN) to generate counterfactual explanations for time series classifiers. The proposed technique adapts existing training instances by copying segments from other instances that belong to the desired class.  More recently, MAPIC, an interpretable model for TSC
based on Matrix Profile, shapelets, and decision tree has been proposed by \cite{guidotti2021matrix}. Finally, a counterfactual solution for multivariate time series data called CoMTE has been proposed by Etes et al. \cite{ates2021counterfactual}, which focuses on observing the effect of turning off one variable at a time. 

\section{Preliminary Concepts}
\textbf{Notation.} \textit{ We define a time series $T = \{t_1, t_2, ..., t_m\}$ as an ordered set of real values, where m is the length, then we can define a time series dataset $\textbf{T}=\{T_{0},T_{1},...,T_{n}\}$ as a collection of such time series where each time series has mapped to a mutually exclusive set of classes $C=\{c_1, c_2, ..., c_l\}$ and used to train a time series classifier $f: \textbf{T}\xrightarrow{}C$, where $\textbf{T} \in \mathbb{R}^k$ is the $k$-dimensional feature space. For each $T_i$ in time series dataset $\textbf{T}$ associated with a class $f(T_i)=c_{i}$, a counterfactual explanation model $\mathcal{M}$ generates a perturbed sample $T'_i$ with the minimal perturbation that leads to $f(T'_i)=c'_i$ such that $c_i \neq c'_i$.} 

According to \cite{looveren2021interpretable} and \cite{delaney2021instance}, a desire counterfactual instance $x_{cf}$ should obey the following initial properties :

\begin{compactitem}
    \item \textbf{Validity}: The prediction of the to-be-explained model $f$ on the counterfactual instance $x_{cf}$ needs to be different from the prediction of the to-be-explained model $f$ on the original instance $x$ (i.e., if $f(x)=c_i$ and $f(x_{cf})=c_j$, then $c_i \neq c_j$).

    \item \textbf{Proximity}: The to-be-explained query needs to be close to the generated counterfactual instance, which means the distance between $x$ and $x_{cf}$ should be minimal.
    
    \item \textbf{Sparsity}: The perturbation $\delta$ changing the original instance $x_0$ into $x_{cf} = x_0 + \delta$ should be sparse (i.e., $\delta \approx \epsilon$).
    
    \item \textbf{Model-agnosticism}: The counterfactual explanation model should produce a solution independent of the classification model $f$, high-quality counterfactuals without prior knowledge of the gradient values derived from optimization-based classification models should be generated. 
    
    \item \textbf{Contiguity}: The counterfactual instance $x_{cf} = x + \delta$ needs to be perturbed in a single contiguous segment which makes the solution semantically meaningful.
    
    \item  \textbf{Efficiency}: The counterfactual instance $x_{cf}$ needs to be found fast enough to ensure it can be used in a real-life setting.

\end{compactitem}

\section{Motif-Guided Counterfactual Explanation (MG-CF)}
\begin{figure*}[h!]
\centering
\includegraphics[scale = .5]{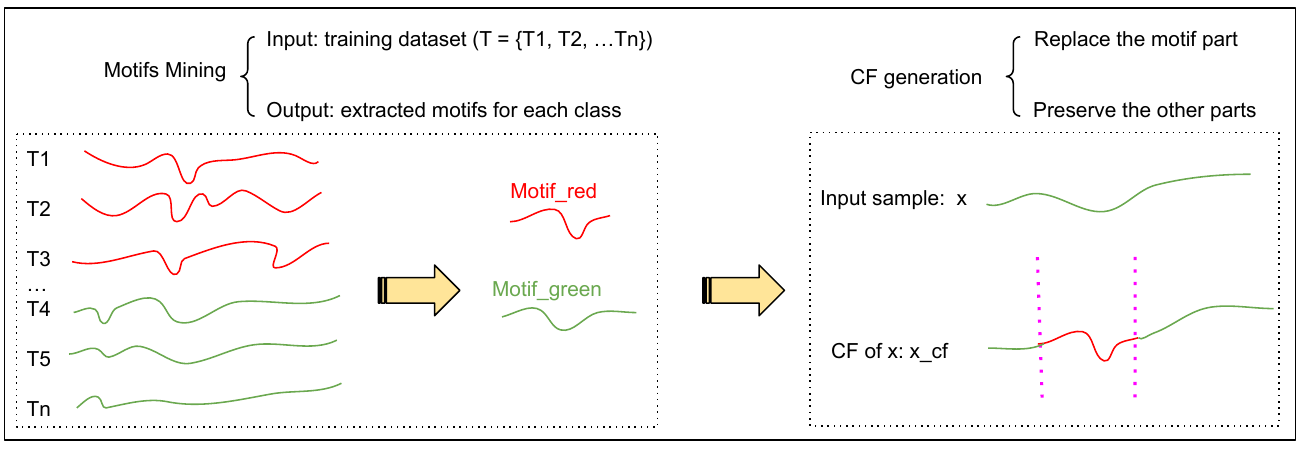}
\caption{Motif-Guided Counterfactual Explanation}
\label{Fig: MGcf}
\label{Fig:archi}
\end{figure*}

In this section, we describe our proposed Motif-Guided Counterfactual Explanation method. The general architecture is illustrated in Figure \ref{Fig: MGcf}, there are two main steps: motifs mining and counterfactual explanation generation. 

\subsection{Motif mining}
Motifs are time series sub-sequences that are maximally representative of a class \cite{ye2009time}. Motifs provide interpretable information beneficial for domain experts to better understand their data. To mine the most prominent motifs, we apply the Shapelet Transform (ST) algorithm proposed by \cite{lines2012shapelet}. To generate the most prominent motifs, a set of candidate motifs need to be generated, a distance measure between a motif and each time series instance should be defined, and a measure of the discriminatory power of the mined motifs \cite{lines2012shapelet} need to be defined. The details of the algorithm is shown in Algorithm \ref{alg:1}. The algorithm starts by sampling specific-length intervals from the time series instance to generate candidate motifs (line 7). The motifs' discriminatory power is then estimated by first computing their distance to all the instances from each class (line 9), and then computing the distance correlation to the target class (line 10). Finally, the motifs are sorted and only the best motif from each class is retained (lines 14, 16). Fig \ref{fig:shapelet0} and \ref{fig:shapelet1} show the two extracted motifs of different classes from ECG 200 dataset. 

\subsubsection{Candidates generation}
Given a time series $T$ with length m, a subsequence $S$ of length $l$ of time series $T$ is a contiguous sequence of $l$ points in $T$, where $l$ $\leq$ $m$. For a given m-lengthed time series, there are $(m-l) + 1$ distinct subsequences of length $l$. The set of all subsequences of length $l$ for a dataset is described as:  
\begin{equation}
    W_l = \{W_{1,l},...,W_{n,l}\},
\end{equation}
where $W_{i,l}$ is defined as the set of all subsequences of length l for time series $T_i$.
To speed up the process of finding motifs, instead of considering all the lengths that range in (3, m), similar to \cite{lines2012shapelet}, we only consider 0.3, 0.5 and 0.7 percent length of the whole time series for each dataset. In this case, the set of all candidate motifs for data set $T$ is:
\begin{equation}
    W = \{W_{0.3m}, W_{0.5m},W_{0.7m}\},
\end{equation}
We denote the candidates generation as \textit{generateCandidates} in Algorithm 1. 
\subsubsection{Measuring Distances}
The distance between two subsequences $S$ and $R$ of length $l$ is defined as
\begin{equation}
    dist(S, R) = \sum_{i=1}^{l}(s_i - r_i)^2
\end{equation}
To calculate all distances between a candidate motif $S$ and all series in T, we denote $d_{i,s}$ as the distance between a subsequence $S$ of $l$ and time series $T_i$, then
\begin{equation}
    d_{i,S} = \mathop{min}_{R \in W_{i,l}}dist(S,R)
\end{equation}
\begin{equation} \label{eq:ds}
    D_S = <d_{1,S}, d_{2,S}, ..., d_{n,S}>
\end{equation}
We denote the process of measuring distance as \textit{findDistances} in Algorithm 1. 
\subsubsection{Measuring the discriminatory power of a motif}
Information gain has been used to evaluate the quality of a motif in the previous shapelet papers \cite{ye2009time}, \cite{ye2011time}, \cite{mueen2011logical}. Firstly, the distance list $D_S$, defined in Equation~\ref{eq:ds}, is sorted, then the information gain is evaluated on the class values for each possible split value. Ye et al. propose to use an early abandon technique on calculating $D_S$ through maintaining an upper bound on the quality of the candidate whilst generating $D_S$ \cite{ye2011time}. If the upper bound falls below the best found so far, the calculation of $D_S$ can be abandoned. After each $d_{i,S}$ is found, the current best information gain split is calculated and the upper bound is found, assuming the most optimistic division of the remaining distances. We denote the process of measuring the discriminatory power of a motif as \textit{assessCandidate} in Algorithm 1. 

\begin{algorithm}[h!]
\caption{Motifs mining} 
\label{alg:1}
\textbf{Input:} 
Training set samples $\textbf{T}$ with labeled binary classes C = [0, 1]\\
\textbf{Output:} 
Extracted motifs for each class
\begin{algorithmic}[1]
\State Motifs = $\emptyset$
\State N = length($\textbf{T}$[0])
\Comment{Number of time series samples}
\State m = length($\textbf{T}$[0][0]) 
\Comment{Length of time series}
\For{$\textbf{T}_i$ $\leftarrow$ $\textbf{T}_1$ to $\textbf{T}_N$}
    \State Motifs $\gets$ $\emptyset$
    \For{l in [0.3m, 0.5m, 0.7m]}
         \State $W_{i,l}$  $\gets$ generateCandidates($\textbf{T}_i$, l)
         \For{all subsequences S in $W_{i,l}$}
             \State $D_S$ $\gets$ findDistances(S, $W_{i,l}$)
             \State quality $\gets$ assessCandidate(S, $D_S$)
             \State Motifs.add(i,start\_idx, end\_idx, S, quality)
             \Comment{The index of time series, the start idx and end idx of motifs will be stored}
         \EndFor
    \EndFor
\State sortByQuality(Motifs)
\EndFor
\State \Return $[[i\_0, start\_idx\_0, start\_idx\_0] , [i\_1, start\_idx\_1, start\_idx\_1]]$
\Comment{return the index information for motifs of different classes }
\end{algorithmic}
\end{algorithm}

\subsection{CF generation}
\begin{figure*}[h!]
        \centering
        \begin{subfigure}[b]{0.43\textwidth}
            \centering
            \includegraphics[width=\textwidth]{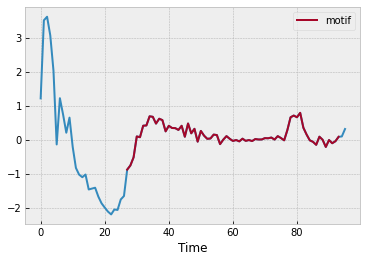}
            \caption[]%
            {{\small Myocardial Infarction Class}}   
            \label{fig:shapelet0}
        \end{subfigure}
        % \hfill
        \begin{subfigure}[b]{0.43\textwidth}  
            \centering 
            \includegraphics[width=\textwidth]{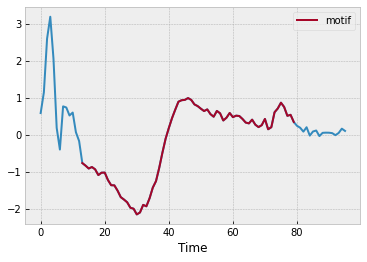}
            \caption[]%
            {{\small Normal Heartbeat class}}    
            \label{fig:shapelet1}
        \end{subfigure}
      
        % \vskip\baselineskip
        \begin{subfigure}[b]{0.43\textwidth}   
            \centering 
            \includegraphics[width=\textwidth]{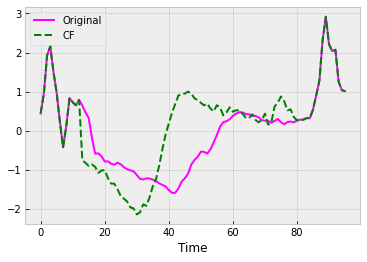}
            \caption[]%
            {{\small Myocardial Infarction query and CF }}    
            \label{fig:cf1}
        \end{subfigure}
        % \hfill
        \begin{subfigure}[b]{0.43\textwidth}   
            \centering 
            \includegraphics[width=\textwidth]{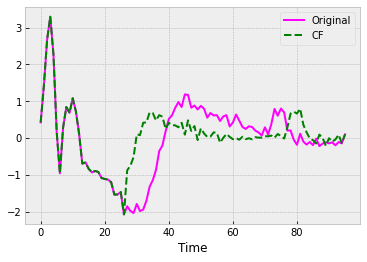}
            \caption[]%
            {{\small Normal Heartbeat query and CF }}    
            \label{fig:cf2}
        \end{subfigure}
        \caption[]%
        {\small Example MG-CF explanations for the ECG200 dataset} 
        \label{fig:cfexample}
\end{figure*}

After motifs are extracted from the training dataset, the next step is to generate CF explanations. In this section, we introduce the counterfactual explanation generation process as outlined at Algorithm \ref{alg:2}. The CF generation starts by using the pre-trained model $f$ to predict the classification result of the test dataset $\textit{samples}$ (line 1). We set the target classes of the test dataset as the opposite label of the prediction results (line 2). To generate an explanation of a query time series $X$, we consider the motif's existing segment from the target class as the most prominent segment (lines 4-7). Therefore, we replace the motif existing part with the target class motif and preserve the remaining parts (line 9). In this case, the counterfactual explanation for the input query time series is generated (line 10).  It is important to note that MG-CF does not require the use of class activation maps which makes it a model-agnostic model. Figure \ref{fig:cfexample} shows examples of generated counterfactual explanations for the ECG 200 dataset of the Myocardial Infarction and Normal Heartbeat class.  

\begin{algorithm}[h!]
\caption{CF generation} 
\label{alg:2}
\textbf{Input:} 
Test set samples $\textit{samples}$, pretrainined time series classifier $f$, extracted motifs for each class from Motif mining algorithm

\textbf{Output:} $\textbf{CF}$, Counterfactual explanation for each sample in test set
\begin{algorithmic}[1]
\State preds = $f$.predict(\textit{samples})
\Comment{Applying the test data on f to get the prediction results}
\State targets= the opposite label of the preds
\Comment{Get the target labels based on the prediction results}
\For{$\textit{sample}$ $\leftarrow$ $\textit{samples}$}
    \If{target(sample) == 0}
        \State i, start\_idx, end\_idx = [i\_0, start\_idx\_0, end\_idx\_0]
        \Comment{extracted index information of the motif from class 0 }
    \Else
        \State i, start\_idx, end\_idx = [i\_1, start\_idx\_1, end\_idx\_1]
        \Comment{extracted index information of the motif from class 1 }
    \EndIf
    \State sample[start\_idx, end\_idx] = T[i][start\_idx, end\_idx]
    \State cf\_sample = sample
    \State CF.append(cf\_sample)
\EndFor
\State \Return $\textbf{CF}$
\end{algorithmic}
\end{algorithm}

% % \raggedbottom
% \vspace{-10pt}
\section{Experimental Evaluation}
In this section, we outline our experimental design and discuss our findings. We conduct our experiments on the publicly-available univariate time series data sets from the University of California at Riverside (UCR) Time Series Classification Archive \cite{dau2019ucr}. We test our model on five real-life datasets from various domains (image, spectro, ECG, and motion).
The number of labels of each dataset is binary and the time series classes are evenly sampled across the training and testing partitions. The length of the time series in the data sets varies between 96 and 512. Table~\ref{tab:UCR} shows all the dataset details (number of classes, time series length, training size, testing size, and the type of time series). 
\subsection{Baseline Methods}
We evaluate our proposed model with two other baselines, namely, Alibi \cite{JMLR:v22:21-0017} and Native guide counterfactual explainers \cite{delaney2021instance}.
\begin{itemize}
    \item \textbf{Alibi Counterfactual (Alibi): } The Alibi generates counterfactuals by optimizing a new objective function that includes class prototypes generated from encoders \cite{JMLR:v22:21-0017}. The method leverages the resulting autoencoder reconstruction error after decoding the counterfactual instance $x_{cf}$ as a measure of fitness to the training data distribution. The additional loss term $L_{AU}$ penalizes out-of-distribution counterfactual instances.
    
    \item \textbf{ Native guide counterfactual (NG-CF):} NG-CF is the latest time series counterfactual method that perturbs the query time series by inserting segments from the opposing class \cite{delaney2021instance}. NG-CF uses Dynamic Barycenter (DBA) averaging of the query time series $x$ and the nearest unlike neighbor from another class to generate the counterfactual example.
\end{itemize}

The source code of our model, competing baselines, and the experimental dataset are available on the MG-CF project website \footnote{\url{https://sites.google.com/view/naive-shapelets-guided-cf}}.

\subsection{Experimental Results \label{Sec:ExpResults}}
The goal of our experiments is to assess the performance of the baseline methods with respect to all the desired properties of an ideal counterfactual method. To evaluate our proposed method, we compare our method with the other two baselines in terms of several metrics: L1 distance (proximity), sparsity level, number of independent segments is perturbing (contiguity), the label flip rate (validity), and the run time (efficiency).

L1 distance, which is defined in Equation~\ref{equ:L1}, measures the distance between the counterfactuals and the original samples, a smaller L1 distance is desired.

\begin{equation}
    proximity = \| x_{cf} - x\|
    \label{equ:L1} 
\end{equation}

\begin{figure}[!h]
     \centering
     \begin{subfigure}[b]{0.49\textwidth}
         \centering
         \includegraphics[width=\textwidth]{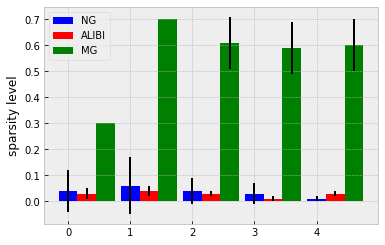}
        %  \caption{$y=x$}
         \label{fig:sparsity}
     \end{subfigure}
     \hfill
     \begin{subfigure}[b]{0.49\textwidth}
         \centering
         \includegraphics[width=\textwidth]{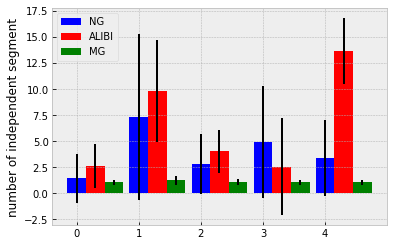}
        %  \caption{$y=3sinx$}
         \label{fig:segments}
     \end{subfigure}
     \caption{The sparsity level(the higher the better) and number of independent segments(the lower the better) of the CF explanations}
     \label{fig:sparsity_segments}
\end{figure}

\begin{table*}[!t]
\centering
\caption{UCR datasets metadata (each dataset is binary labeled)}
\begin{tabular}{|c||c|c|c|c|c|c}
\hline
\label{tab:UCR}
ID & Dataset Name   & TS length & DS train size & DS test size & Type\\ \hline
0  & ECG200         & 96        & 100   &100  &ECG\\ \hline
1  & Coffee         & 286       & 28    &28   &SPECTRO\\ \hline
2  & GunPoint       & 150       & 50    &150  &MOTION\\ \hline
3  & BeetleFly      & 470       & 20    &20   &IMAGE\\ \hline
4  & BirdChicken    & 512       & 20    &20   &IMAGE\\ 
\hline
\end{tabular}
\end{table*}
Sparsity level, which indicates the level of time series perturbations. A high sparsity level that is approaching $100\%$ is desirable, which means the time series perturbations made in $x$ to achieve $x_{cf}$ is minimal. We computed the sparsity level using a new metric that we formulate in Equations~\ref{equ:spar}-\ref{equ:func}.
\begin{equation}\label{equ:spar}
    sparsity = 1- \frac{\sum_{i=0}^{len(x)}g(x_{cf_i}, x_i)}{len(x)} 
\end{equation}
\begin{equation}\label{equ:func}
    g(x, y)=\left\{\begin{array}{ll}
    1, & \text { if } x \neq y \\
    0, & \text { otherwise }
    \end{array}\right.
\end{equation}
The number of independent non-contiguous segments is also investigated to show the contiguity, which is shown in Fig \ref{fig:sparsity_segments}. The lower the number of independent non-contiguous segments the better. Finally, we define the validity metric by comparing the label flip rate for the prediction of the counterfactual explanation result, we computed the flip rate following the formula of Equation~\ref{equ:flipr}.
\begin{equation}\label{equ:flipr}
    flip\_rate = \frac{num\_flipped}{num\_testsample},
\end{equation}
where we denote the num\_flipped as the number of generated counterfactual explanation samples that flip the label, and the num\_testsample is the total number of samples in the test dataset that we used to generate counterfactual explanation samples. The closer the label flip rate is to 1, the better. 

%  Figures \ref{fig:sparsity_segments} and \ref{fig:runtimeL1fliprate} show the results of all the datasets evaluated over the different metrics. 

\begin{figure}[!h]
     \centering
     \begin{subfigure}[b]{0.45\textwidth}
         \centering
         \includegraphics[width=\textwidth]{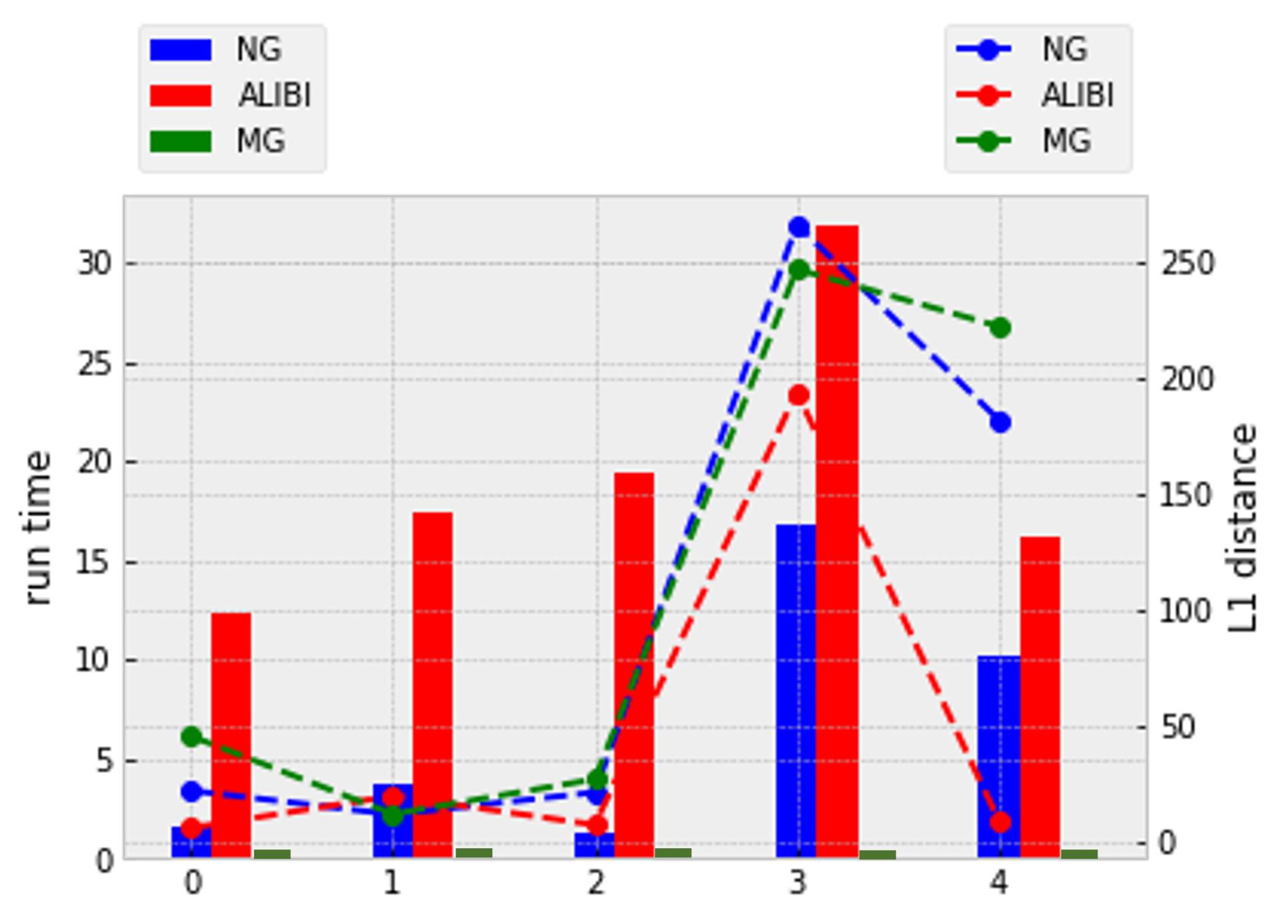}
        %  \caption{$y=x$}
         \label{fig:runtimeL1}
     \end{subfigure}
     \hfill
     \begin{subfigure}[b]{0.45\textwidth}
         \centering
         \includegraphics[width=\textwidth]{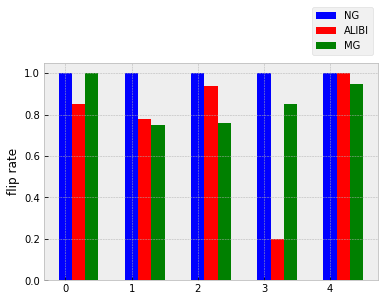}
        %  \caption{$y=3sinx$}
         \label{fig:fliprate}
     \end{subfigure}
     \caption{The average running time (the lower the better), average L1 distance (the lower the better), and the label flip rate (the higher the better) of the CF explanations}
     \label{fig:runtimeL1fliprate}
\end{figure}

Fig \ref{fig:sparsity_segments} and \ref{fig:runtimeL1fliprate} show the evaluation results on the CF desired properties assessed using the aforementioned metrics. Since our proposed method relies on an instance-based counterfactual explanation, which generates a CF for each time series in the dataset, we calculate the mean value and the standard deviation for all CF samples for each dataset to show the overall performances. We note that Alibi achieves the lowest L1 distance (proximity) in the four datasets, which is highly desirable. However, Alibi minimizes proximity in the cost of sparsity level, contiguity (number of independent segments), and efficiency (running time). Alibi achieves the lowest sparsity level and run time, which entails that the method generates CF explanations with less proximity, contiguity, and efficiency. 

We also note that NG-CF results in highly valid counterfactual explanations that are guaranteed to have the desired class. Since the method does not rely on any optimization function to generate new shapes, it simply copies fragments from existing training time series instances that the prediction model $f$ has already learned. Therefore, it is expected that the prediction model $f$ recognizes the copied segments with high confidence. Although highly valid, NG-CF sacrifices the other metrics (sparsity, contiguity) for maximizing validity. Figures \ref{fig:sparsity_segments} and \ref{fig:runtimeL1fliprate} show that our proposed MG-CF results in a higher sparsity level, a lower number of independent perturbed segments (contiguity), and a lower running time (efficiency) compared to NG-CF and Alibi. Finally, although Alibi and NG-CF can lead to high proximity and validity counterfactuals, they are not valid and sparse respectively. Thus, MG-CF shows a good balance across all the metrics without maximizing one property and compromising the other. 
 \vspace{-10pt}
 
\section{Conclusion}
In this paper, we propose a novel model that derives intuitive, interpretable post-hoc counterfactual explanations of time series classification models that finds balance between validity, proximity, sparsity, contiguity, and efficiency desirable properties. The main limitation of current time series explainers is that they produce perturbations that maximize a metric on the cost of others. This shortcoming limits the interpretability of their resulting counterfactual explanations. We address these challenges by proposing MG-CF, a motif-based model that produces high-quality counterfactual explanations that are contiguous and sparse. The MG-CF method guides the perturbations on the query time series resulting in significantly sparse and more contiguous explanations than other state-of-the-art methods. This is the first effort to leverage the a-priori mined motifs to produce high-quality counterfactual explanations. There are spaces for extensions of our work with time series datasets with high dimensionality and multiple-classes case. As a future direction of this work, we would like to extend our work to multiple-class case time series dataset and also focus on generating the counterfactual explanations for high dimensional time series data.

\bibliographystyle{splncs04}
\bibliography{mybib}


% LNCS Homepage, \url{http://www.springer.com/lncs}. Last accessed 4
\end{document}